\documentclass{article}
\usepackage{spconf,amsmath,epsfig}

\usepackage{cite}
\usepackage{url}
\usepackage{lipsum}
\usepackage{xurl}
\usepackage{siunitx}
\usepackage{pifont}
\usepackage{graphicx}
\usepackage{caption}
\usepackage{subcaption}
\usepackage{upgreek}
\usepackage{amsfonts}
\usepackage{makecell}
\usepackage{multirow}
\usepackage{booktabs}
\usepackage{tabularx}
\usepackage{cleveref}
\crefname{Fig}{Figure}{Figure}
\usepackage[usenames,dvipsnames]{xcolor} 
\usepackage{colortbl}
\usepackage{algorithm}
\usepackage{pgffor}
\usepackage{algorithmicx}
\usepackage{algpseudocode} %
\usepackage[export]{adjustbox}
\usepackage[short]{optidef}
\usepackage{soul}
\usepackage{ifthen}
\usepackage{tikz}
\usepackage{pgfplots}
\usepackage{pgfplotstable}
\usepackage{mathtools}
\usepgfplotslibrary{fillbetween}

\title{Active Learning Guided Fine-Tuning for enhancing Self-Supervised Based Multi-Label Classification of Remote Sensing Images}
\name{Lars Möllenbrok{\normalfont\textsuperscript{1,2}} and Beg\"{u}m Demir{\normalfont\textsuperscript{1,2}}}
\address{\textsuperscript{1}Faculty of Electrical Engineering and Computer Science, Technische Universit\"at Berlin, Germany\\
\textsuperscript{2}BIFOLD - Berlin Institute for the Foundations of Learning and Data, Germany}

\begin{document}
%
\maketitle
\begin{abstract}
In recent years, deep neural networks (DNNs) have been found very successful for multi-label classification (MLC) of remote sensing (RS) images. Self-supervised pre-training combined with fine-tuning on a randomly selected small training set has become a popular approach to minimize annotation efforts of data-demanding DNNs. However, fine-tuning on a small and biased training set may limit model performance. To address this issue, we investigate the effectiveness of the joint use of self-supervised pre-training with active learning (AL). The considered AL strategy aims at guiding the MLC fine-tuning of a self-supervised model by selecting informative training samples to annotate in an iterative manner. Experimental results show the effectiveness of applying AL-guided fine-tuning (particularly for the case where strong class-imbalance is present in MLC problems) compared to the application of fine-tuning using a randomly constructed small training set.
\end{abstract}


\section{Introduction}
\noindent
Automatically assigning multiple land-use land-cover class labels (i.e., multi-labels) to image scenes from an archive of remote sensing (RS) images has become an important task in RS. Therefore, developing accurate multi-label image scene classification (MLC) methods is a growing research interest in RS.
Deep learning (DL) based methods have recently seen a rise in popularity in the context of MLC problems, as they can learn rich features to describe the complex spectral and spatial content of RS images  \cite{shendryk2018deep, sumbul2020deep}. For example, in \cite{8633359} a convolutional neural network is trained using an MLC-adjusted data-augmentation pipeline. To train DL models in a supervised setting a large number of images (i.e., samples) that are annotated with high-quality multi-labels is needed. However, collecting multi-labels is time-consuming, complex and costly in operational scenarios \cite{zegeye2018novel}. To address this issue self-supervised learning (SSL) methods can be used in the context of learning general image features from unlabeled data without using any human-annotated labels by considering auxiliary learning objectives. For example, in \cite{zbontar2021barlow} image representations are learned by forcing the cross-correlation matrix of the outputs of a siamese-style architecture, which is fed with augmented views of the same image, to be close to the identity matrix. After self-supervised training of a model, the learned parameters serve as a pre-trained model and can be transferred to the MLC task by fine-tuning on a smaller set of samples annotated with multi-labels\cite{9875399}. These samples for fine-tuning are usually obtained by annotating a random subset of the available unlabeled data, where the amount of selected samples depends on the annotation budget (i.e. the number of images that can be afforded to label) and the desired performance (a higher annotation budget typically results in better performance). However, randomly selecting samples to be labeled may result in a biased set for fine-tuning, and therefore may reduce performance of the classification model. In MLC scenarios, where class occurrences often follow long-tail distributions, this issue is highly critical.

Active learning (AL) on the other hand has been found successful for finding highly informative samples from a set of unlabeled samples that, when annotated and added to the training set, can improve model performance significantly\cite{mollenbrok2022deep}. Generally, the informativeness of a sample is assessed by jointly evaluating two criteria: i) uncertainty and ii) diversity. The uncertainty criterion measures the confidence of the model to correctly assign label to a given sample, whereas the diversity criterion ensures that the selected samples are diverse to each other. For example, in \cite{ash2019deep} uncertainty and diversity are jointly assessed by sampling loss gradient approximations that are as distant as possible to each other based on a greedy strategy. During each AL iteration, the most informative samples are selected and annotated by a human expert and added to the labeled training set. The process is terminated when either performance converges or the budget limit for labeling is reached. While AL is successfully employed for enriching the training set, it does not make use of available unlabeled data for the training.

In this paper, we investigate the effectiveness of joint use of self-supervised pre-training and AL in MLC problems. To this end, we enhance a self-supervised model through AL-guided fine-tuning. 
The benefits of combining SSL methods for pre-training with AL-guided fine-tuning are: i) learning parameters for initialising the model from unlabeled data, ii) obtaining an optimized set of labeled samples that are highly informative for fine-tuning the model, while avoiding labeling non-informative samples; and iii) achieving stronger implicit robustness to class imbalances in the archive.

\section{Methodology}
\noindent
Let $\mathcal{X}= \{\mathbf{X}_1,...,\mathbf{X}_N\}$ be an archive that contains $N$ RS images, where $\mathbf{X}_i$ is the $i$-th image. We assume that a small and biased set of labeled images $\mathcal{T}^1=\{(\mathbf{X}^l_1,\mathbf{y}_1),...,(\mathbf{X}^l_M,\mathbf{y}_M)\}$ is available. For an image $\mathbf{X}^l_i$ the multi-label vector $\mathbf{y}_i = [y_1^{(i)},...,y^{(i)}_C] \in \{0,1\}^C$ indicates the presence or absence of each of $C$ unique classes. Typically $\mathcal{T}^1$ is much smaller than the archive $\mathcal{X}$ (i.e. $N \gg M$) and might be biased. We aim to learn the parameters for a multi-label classification model $\mathcal{F}$ under a labeling budget $B$. To this end, we follow an approach that consists of two steps: i) self-supervised pre-training on the archive set $\mathcal{X}$ for model initialisation; and ii) AL-guided fine-tuning of the classification model on an iteratively enriched labeled training set.


\subsection{Self-supervised pre-training for model initialisation}
\noindent
In the first step, we pre-train the model in self-supervised manner on the archive set $\mathcal{X}$. Let the model $\mathcal{F}=g \circ f$ be composed of a convolutional backbone $f$ followed by a classification head $g$. To find initial parameters for the convolutional backbone $f$, we make use of the BYOL \cite{grill2020bootstrap} framework. In this framework, $f$ is embedded as an encoder into an asymmetric two-branch architecture consisting of an online network and a target network. Both the online and the target network consist of an encoder followed by a projector. The online network additionally employs a predictor $q$ on top of its projector. For two distributions of image augmentations $\mathcal{A}$ and $\mathcal{A}'$, and given an image $\mathbf{X}_i$, online network and target network receive two different augmented views of the image $a(\mathbf{X}_i)$ and $a'(\mathbf{X}_i)$ as inputs, where $a \sim \mathcal{A}$ and $a' \sim \mathcal{A}' $. The online network is trained to predict the output of the target network by minimizing the norm-adjusted mean squared error (MSE) loss defined as follows:
\begin{equation}
    \mathcal{L}_{MSE}(\mathbf{X}_i) = \left \Vert \frac{q(\mathbf{z}^{(i)}_{on})}{\|q(\mathbf{z}^{(i)}_{on})\|_2} - \frac{\mathbf{z}^{(i)}_{tar}}{\|\mathbf{z}^{(i)}_{tar}\|_2} \right \Vert ^2_2
\end{equation}
where $\mathbf{z}^{(i)}_{on}$ and $\mathbf{z}^{(i)}_{tar}$ are the outputs of the projectors for image $\mathbf{X}_i$ of the online network and the target network, respectively.
The target network is not trained, but instead it is updated as the exponential moving average of the online network parameters.

For a more detailed description about the self-supervised training pipeline and information about the distributions of image augmentations $\mathcal{A}$ and $\mathcal{A'}$, the reader is referred to \cite{grill2020bootstrap}.

\subsection{AL-guided fine-tuning}
\noindent
In the second step, we iteratively fine-tune the model $\mathcal{F}$ on the labeled set that is enriched using an AL strategy.
 At each iteration $t$, starting from $t=1$, we fine-tune $\mathcal{F}$ on $\mathcal{T}^{t}$ using binary cross-entropy loss defined as:
 \begin{equation}
     \mathcal{L}_{BCE}(\mathbf{X}_i,\mathbf{y}_i) \! = \! - \frac{1}{C} \! \sum^C_{j=1} y^{(i)}_j \log( p^{(i)}_j) +  (1-y^{(i)}_j) \log(1-p^{(i)}_j)
 \end{equation}
 where $p^{(i)}_j$ is the probability for the $j$-th class being present in image $\mathbf{X}_i$. After fine-tuning the model $\mathcal{F}$ on the labeled samples, we aim to find the $b$ (which corresponds to the labeling budget per iteration) most informative samples from the unlabeled set $\mathcal{U}^{t} = \mathcal{X} \setminus \mathcal{T}^{t}$ to enrich the training set. 

 To select the most informative samples we make use of Measuring Magnitude of Approximated Gradient Embeddings + Clustering (MGE+Clustering) \cite{mollenbrok2022deep} query function. It assesses uncertainty based on magnitude of approximated gradient embeddings (MGE) and achieves sample diversity using a clustering approach.

In MGE, the uncertainty of a sample is estimated by the means of loss gradients. The magnitude of the gradient determines the extend to which the network parameters change. Therefore, samples with a large gradient can be seen as more uncertain and to have more impact to improve model performance during training, since they result in larger changes of model parameters. Since the loss gradients cannot be computed for an unlabeled sample directly (knowledge of the multi-label would be required), a gradient approximation is used to estimate the uncertainty of a sample. The gradient approximations are computed with respect to the weights of the penultimate layer based on the pseudo-labels. Let $\hat{\mathbf{y}}_i$ be the pseudo-labels for an image $\mathbf{X}_i$ obtained by using a $0.5$ threshold on probabilities predicted by the model. The approximation of the gradient embeddings with respect to the weights $W$ of the last layer of $\mathcal{F}$ for the loss $\mathcal{L}_{BCE}$ is given by 
\begin{equation}
g_{\mathbf{X}}^{\hat{\mathbf{y}}} = \nabla_W \mathcal{L}_{BCE}(\mathbf{X}_i,\hat{\mathbf{y}}_i).
\end{equation}
For each unlabeled sample the gradient approximation is initially computed and then its magnitude $\|g_{\mathbf{X}}^{\hat{\mathbf{y}}}\|_2$ is determined to assess the uncertainty of the sample (and thus estimate its impact for training the classification model).

After measuring the uncertainty of the unlabeled samples from $\mathcal{U}^{t}$, MGE+Clustering \cite{mollenbrok2022deep} ensures the diversity of selected samples using a clustering approach. Clustering has been found effective in ensuring diversity, since samples from different clusters are implicitly sparse in the feature space. The learned features of the $m > b$ most uncertain samples (i.e. the samples with the largest gradient approximations) are clustered into $b$ different clusters using Kmeans++ \cite{arthur2007k} algorithm. Then, from each cluster the most uncertain sample is selected for labeling, resulting in $b$ selected samples. Due to the selection of one sample from each cluster, the diversity of samples at each AL iteration is achieved.

Once the $b$ most informative samples are selected, they are labeled by a human expert and the newly labeled samples are added to $\mathcal{T}^{t}$ to form the new training set $\mathcal{T}^{t + 1}$. The AL iterations are repeated until the total labeling budget $B$ is spent.
Note that during the iterative AL-guided fine-tuning, at each iteration the model is re-initialized with the parameters learned in the fine-tuning step of the previous iteration.

\section{Experimental Results}
\noindent
Experiments were conducted on the UCMerced \cite{6257473} dataset with multi-labels obtained from \cite{8089668}. It contains 2100 RGB images of size 256 $\times$ 256 pixels with 30cm spatial resolution. In total, there are 17 unique classes and the number of classes associated to a single image varies between 1 and 8. Class occurrences range from 100 to 1331. We randomly divided the images into a validation set of 525 samples, a test set of 525 samples and a pool set containing 1050 samples. We used DenseNet121 as the architecture for the classification model. For self-supervised pre-training we set the batch size to 100 and trained for 1000 epochs with a learning rate of 0.001 using ADAM optimizer. For AL-guided fine-tuning, we randomly chose 40 samples from the pool to form the initial labeled training set. At each iteration, we used a labeling budget of $b=20$. During fine-tuning, we set the batch size to 10 and train for 100 epochs with a learning rate of 0.025 and SGD optimizer. After 80 epochs, we reduced the learning rate by a factor of 0.1. The data augmentation methods that we use for fine-tuning include random horizontal flipping and random rotation of $\{0,90,180,270\}$ degrees. Results are reported in terms of $F_1$ scores and were averaged over five runs.

\begin{figure}
\centering
\subfloat[]{\includegraphics[width=7.5cm]{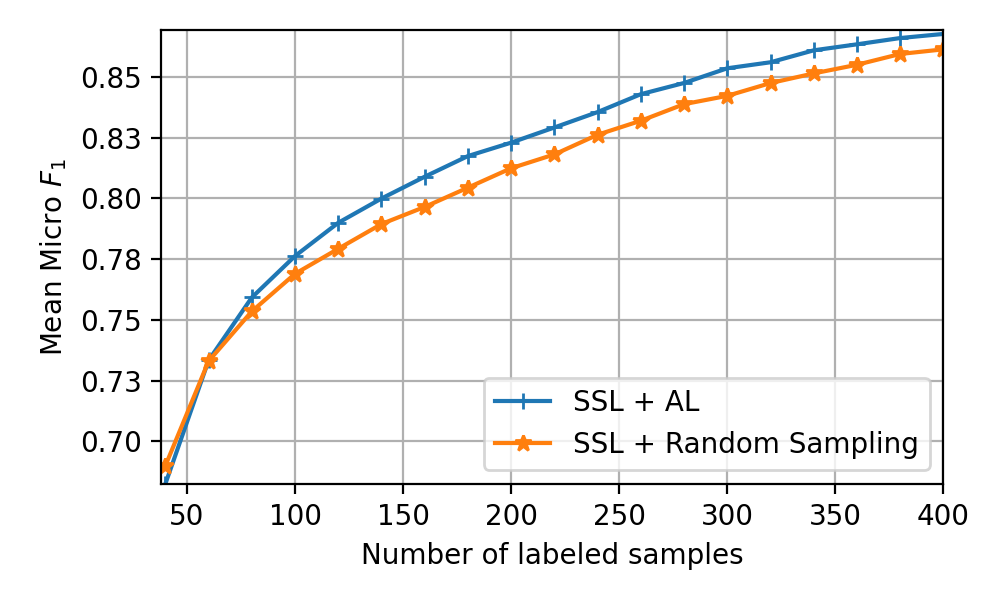}
\vspace*{-2mm}
}
\hfil
\subfloat[]{\includegraphics[width=7.5cm]{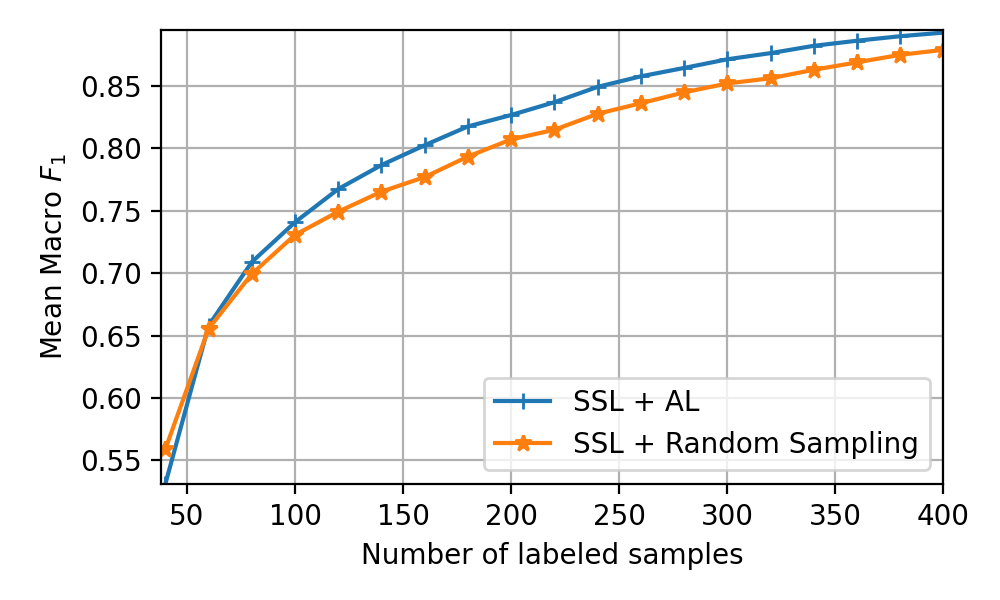}}
\caption{(a) Micro and (b) Macro $F_1$ scores versus the number of labeled samples.}
\label{micro_macro}
\end{figure}

In the first experiment, we compare the performance of AL-guided fine-tuning with that obtained by randomly selecting a set of samples for fine-tuning.
Fig. \ref{micro_macro} shows the performance in terms of $F_1$ scores versus the number of labeled samples. From the figure one can see that using AL-guided fine-tuning complementary to self-supervised pre-training results in better performance. For example, AL-guided fine-tuning provides a micro $F_1$ score of 81.74\%  and a macro $F_1$ score of 81.8\% for 180 labeled samples, whereas random sampling only achieves 80.39\% and 79.8\% in terms of micro and macro $F_1$ scores, respectively. To reach the same performance random sampling needs more than 220 annotated samples. This shows that samples selected through AL are more informative for fine-tuning the model.

\begin{figure}
\centering
\includegraphics[width=7.5cm]{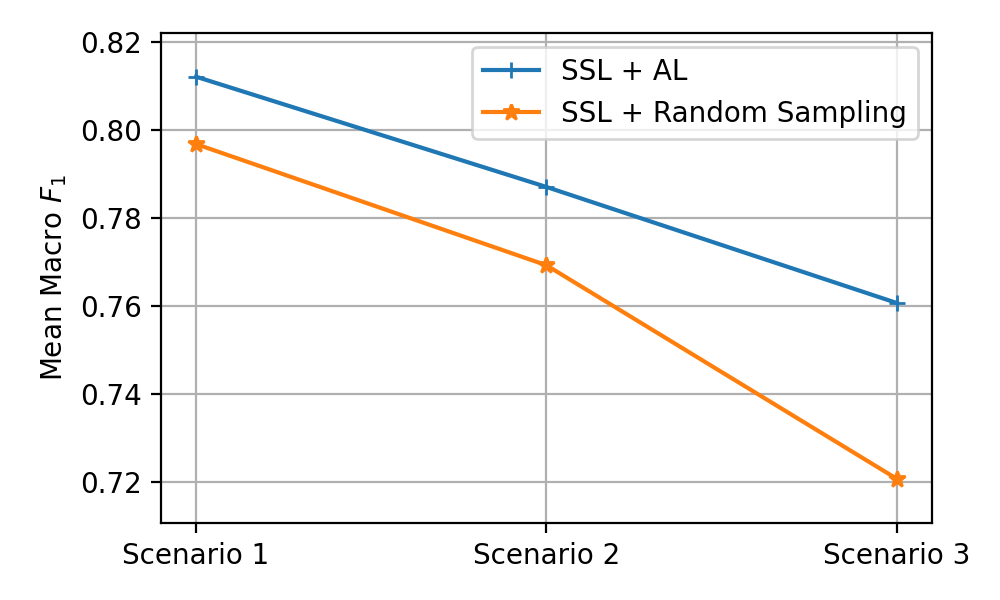}
\caption{Mean Macro $F_1$ scores averaged over 20 iterations in the three different scenarios.}
\label{imbalance}
\end{figure}

In the second experiment we analyse the influence of class imbalance in the archive $\mathcal{X}$ on the performance of the classification model. To this end, we compare three different scenarios associated to different degrees of class imbalance of $\mathcal{X}$. In the Scenario 1, we use the pool set as defined before. In Scenario 2 and Scenario 3, we randomly select three minority classes and randomly remove samples that are associated to these classes. In the Scenario 2, we remove 20 samples from each of the three classes, while in Scenario 3 we remove 40 samples. By this way the class imbalance increases from Scenario 1 to Scenario 3. As minority classes we consider the following classes: Airplane, Chaparral, Court, Dock, Field, Mobilehome, Sea, Ship and Tank. The classes Dock and Ship have 98\% co-occurence. Therefore, we made sure that these classes are not selected together for sample removal. Fig. \ref{imbalance} shows the performance of AL-guided fine-tuning and fine-tuning on a randomly selected training set for the three different scenarios in terms of macro $F_1$ scores averaged over 20 sampling iterations. From the figure, one can see that in all three scenarios AL-guided fine-tuning performs better than fine-tuning on a randomly selected training set. For example, in Scenario 1, AL-guided fine-tuning yields 1.6\% higher average performance in terms of macro $F_1$ score than fine-tuning applied using random sampling. In Scenario 3, AL-guided fine-tuning results in 4\% higher performance in terms of macro $F_1$ score. The divergence of performance over the different scenarios indicates that AL-guided fine-tuning is more robust for the cases of stronger class imbalance. This is due to the fact that the use of AL leads to selection of samples associated to under-represented classes, which is crucial particularly in operational MLC scenarios.

\section{Conclusion}
\noindent
In this paper, we have investigated the effectiveness of the joint use of SSL pre-training and  AL for training an MLC model under a labeling budget. To this end, we use AL to guide the fine-tuning of an SSL pre-trained model for MLC. As an AL method we exploit MGE+Clustering, while for SSL pre-training we use the BYOL framework. The experimental results show that AL-guided fine-tuning results in higher performance than randomly selecting samples for fine-tuning. Furthermore, the results show that AL-guided fine-tuning is more robust to class imbalances.
As future works, we plan to verify our results on larger datasets and to investigate continual learning scenarios, where data distributions can shift over time. 

\section{Acknowledgements}
\noindent
This work is supported by the European Research Council (ERC) through the ERC-2017-STG BigEarth Project under Grant 759764 and by the European Space Agency through the DA4DTE (Demonstrator precursor Digital Assistant interface for Digital Twin Earth) project. The authors would like to thank Gencer Sümbül for his suggestions and comments during this study.



\bibliographystyle{IEEEtran}
\bibliography{refs}

\begin{thebibliography}{10}
\providecommand{\url}[1]{#1}
\csname url@samestyle\endcsname
\providecommand{\newblock}{\relax}
\providecommand{\bibinfo}[2]{#2}
\providecommand{\BIBentrySTDinterwordspacing}{\spaceskip=0pt\relax}
\providecommand{\BIBentryALTinterwordstretchfactor}{4}
\providecommand{\BIBentryALTinterwordspacing}{\spaceskip=\fontdimen2\font plus
\BIBentryALTinterwordstretchfactor\fontdimen3\font minus
  \fontdimen4\font\relax}
\providecommand{\BIBforeignlanguage}[2]{{%
\expandafter\ifx\csname l@#1\endcsname\relax
\typeout{** WARNING: IEEEtran.bst: No hyphenation pattern has been}%
\typeout{** loaded for the language `#1'. Using the pattern for}%
\typeout{** the default language instead.}%
\else
\language=\csname l@#1\endcsname
\fi
#2}}
\providecommand{\BIBdecl}{\relax}
\BIBdecl

\bibitem{shendryk2018deep}
I.~Shendryk, Y.~Rist, R.~Lucas, P.~Thorburn, and C.~Ticehurst, ``Deep
  learning-a new approach for multi-label scene classification in planetscope
  and sentinel-2 imagery,'' \emph{IEEE International Geoscience and Remote
  Sensing Symposium}, pp. 1116--1119, 2018.

\bibitem{sumbul2020deep}
G.~Sumbul and B.~Demir, ``A deep multi-attention driven approach for
  multi-label remote sensing image classification,'' \emph{IEEE Access},
  vol.~8, pp. 95\,934--95\,946, 2020.

\bibitem{8633359}
R.~Stivaktakis, G.~Tsagkatakis, and P.~Tsakalides, ``Deep learning for
  multilabel land cover scene categorization using data augmentation,''
  \emph{IEEE Geoscience and Remote Sensing Letters}, vol.~16, no.~7, pp.
  1031--1035, 2019.

\bibitem{zegeye2018novel}
B.~T. Zegeye and B.~Demir, ``A novel active learning technique for multi-label
  remote sensing image scene classification,'' \emph{SPIE Image and Signal
  Processing for Remote Sensing XXIV}, vol. 10789, pp. 100--107, 2018.

\bibitem{zbontar2021barlow}
J.~Zbontar, L.~Jing, I.~Misra, Y.~LeCun, and S.~Deny, ``Barlow twins:
  Self-supervised learning via redundancy reduction,'' \emph{International
  Conference on Machine Learning}, pp. 12\,310--12\,320, 2021.

\bibitem{9875399}
Y.~Wang, C.~M. Albrecht, N.~A.~A. Braham, L.~Mou, and X.~X. Zhu,
  ``Self-supervised learning in remote sensing: A review,'' \emph{IEEE
  Geoscience and Remote Sensing Magazine}, vol.~10, no.~4, pp. 213--247, 2022.

\bibitem{mollenbrok2022deep}
L.~M{\"o}llenbrok and B.~Demir, ``Deep active learning for multi-label
  classification of remote sensing images,'' \emph{arXiv preprint
  arXiv:2212.01165}, 2022.

\bibitem{ash2019deep}
J.~T. Ash, C.~Zhang, A.~Krishnamurthy, J.~Langford, and A.~Agarwal, ``Deep
  batch active learning by diverse, uncertain gradient lower bounds,''
  \emph{arXiv preprint arXiv:1906.03671}, 2019.

\bibitem{grill2020bootstrap}
J.-B. Grill, F.~Strub, F.~Altch{\'e}, C.~Tallec, P.~Richemond, E.~Buchatskaya,
  C.~Doersch, B.~Avila~Pires, Z.~Guo, M.~Gheshlaghi~Azar \emph{et~al.},
  ``Bootstrap your own latent-a new approach to self-supervised learning,''
  \emph{Advances in Neural Information Processing Systems}, vol.~33, pp.
  21\,271--21\,284, 2020.

\bibitem{arthur2007k}
D.~Arthur and S.~Vassilvitskii, ``K-means++ the advantages of careful
  seeding,'' \emph{Proceedings of the Eighteenth Annual ACM-SIAM Symposium on
  Discrete Algorithms}, pp. 1027--1035, 2007.

\bibitem{6257473}
Y.~Yang and S.~Newsam, ``Geographic image retrieval using local invariant
  features,'' \emph{IEEE Transactions on Geoscience and Remote Sensing},
  vol.~51, no.~2, pp. 818--832, 2013.

\bibitem{8089668}
B.~Chaudhuri, B.~Demir, S.~Chaudhuri, and L.~Bruzzone, ``Multilabel remote
  sensing image retrieval using a semisupervised graph-theoretic method,''
  \emph{IEEE Transactions on Geoscience and Remote Sensing}, vol.~56, no.~2,
  pp. 1144--1158, 2018.

\end{thebibliography}

\end{document}